\documentclass[10pt,conference]{IEEEtran}
\IEEEoverridecommandlockouts
  \usepackage{cite}
  \usepackage[font=small]{caption}
\usepackage{listings}
\usepackage{enumitem}
\usepackage{stfloats}
\usepackage{multirow}
\usepackage{etoolbox}
\makeatletter
\patchcmd{\@makecaption}
  {\scshape}
  {}
  {}
  {}
\makeatother
\usepackage[pdftex]{graphicx}

\usepackage{lipsum}
\usepackage[ruled]{algorithm2e}
\usepackage{amsmath}
\usepackage{acronym}
\usepackage{algorithmic}
\usepackage{array}
\usepackage{mdwmath}
\usepackage{mdwtab}
\usepackage{eqparbox}
\usepackage{subfig}
\usepackage[final]{pdfpages}

\usepackage{xcolor}
\def\BibTeX{{\rm B\kern-.05em{\sc i\kern-.025em b}\kern-.08em
        T\kern-.1667em\lower.7ex\hbox{E}\kern-.125emX}}
\begin{document}

\title{Complexity-aware Adaptive Training and Inference for Edge-Cloud Distributed AI Systems\\
    \thanks{© 2021 IEEE. Personal use of this material is permitted. Permission from IEEE must be obtained for all other uses, in any current or future media, including reprinting/republishing this material for advertising or promotional purposes, creating new
collective works, for resale or redistribution to servers or lists, or reuse of any copyrighted component of this work in other works.}
    \vspace{-1ex}
}
\author{\IEEEauthorblockN{ Yinghan~Long, Indranil Chakraborty, Gopalakrishnan Srinivasan, Kaushik Roy }
\IEEEauthorblockA{\textit{School of Electrical and Computer Engineering, Purdue University} \\
    long273@purdue.edu, ichakra@purdue.edu, srinivg@purdue.edu, kaushik@purdue.edu}
\\[-5.0ex]
}

\maketitle

\begin{abstract}
%Although edge-based smart data processing can be enabled by deploying pretrained models, distributing the model and tasks between edge and cloud can scale and improve machine learning applications. The energy and memory constraints of edge devices would require a complexity-aware model to be adaptively trained with a limited amount of data. 
%and AI are still needed to intelligently adapt to dynamic IoT environments and learn from local data. 
The ubiquitous use of IoT and machine learning applications is creating large amounts of data that require accurate and real-time processing. Although edge-based smart data processing can be enabled by deploying pretrained models, the energy and memory constraints of edge devices necessitate distributed deep learning between the edge and the cloud for complex data. In this paper, we propose a distributed system to exploit both the edge and the cloud for training and inference. We propose a new architecture, MEANet, with a \textit{main block}, an \textit{extension block}, and an \textit{adaptive block} for the edge.
%\textcolor{blue}{The main block identifies instances of easy/hard classes and classifies easy classes with high confidence. The extension and the adaptive blocks adjust the decision boundary between hard classes try to classify the hard instances sent from the main block. If it cannot provide high confidence on such prediction, it is considered complex and sent to the cloud for processing. We also propose an efficient training technique considering the energy constraint of edge devices.} The training technique lends to majority of the inference on edge devices while going to the cloud only for a small set of complex jobs, as determined by the edge network.
% First, the main block is pretrained at the cloud.
%% With what data?
%Second, the main block is deployed in the edge with its parameters fixed, and the other blocks are trained with data belonging to complex classes at the edge.
%% Trained with separate set of complex data?
%Because we reduce the amount of training data and active parameters at the second step,
%% the above expression is not clear -- activate parameters at the second step??? The above steps are not very clear or intuitive. Can we instead state that we propose a distributed network architecture and training technique considering energy constraint of edge devices. The training technique lends to majority of the inference on edge devices while going to the cloud only for a small set of complex jobs.  
%the training cost decreases to around 10-30\%.
%% training cost decreases to ??% or decreases by ??%
The inference process can terminate at either the main block, the extension block, or the cloud. MEANet is trained to categorize inputs into easy/hard/complex classes. The main block identifies instances of easy/hard classes and classifies easy classes with high confidence. Only data with high probabilities of belonging to hard classes would be sent to the extension block for prediction. Further, only if the neural network at the edge shows low confidence in the prediction, the instance is considered complex and sent to the cloud for further processing. The training technique lends to the majority of inference on edge devices while going to the cloud only for a small set of complex jobs. The performance of the proposed system is evaluated via extensive experiments using modified models of ResNets and MobileNetV2 on CIFAR-100 and ImageNet datasets. The results show that the proposed distributed model has improved accuracy and lower energy consumption compared to standard models, indicating its capacity to adapt.
\end{abstract}

% Note that keywords are not normally used for peer review papers.

\section{Introduction}

%IoT and AI; explain the need for edge computing;

% "By moving cognitive computation intense tasks locally on embedded edge devices, not only world-wide internet power consumption growth trend will be reduced, but also users will recover their right to keep their personal data privacy."

The availability of a large amount of data at the edge and the development of Artificial Intelligence (AI) has significantly augmented the paradigm of Internet of Things (IoT). Deep Learning (DL) and deep neural networks (DNN) have achieved remarkable performance in various applications such as computer vision, including object detection, face and visual scene recognition \cite{krizhevsky2012imagenet, he2016deep, CloudtoIOT}. The growth of machine learning applications has, on one hand, positively impacted human lives, but on the other hand, has led to the need to smartly manage the humongous amount of heterogeneous data \cite{CloudtoIOT, IoTML}. Processing this data (which is available at the edge) entirely in the cloud raises concerns about latency, energy, and data privacy. Despite its success, DNNs are computationally expensive, making them unsuitable for resource-constrained edge devices. This necessitates intelligent collaborative decision strategies between the resource constrained edge devices and the cloud.

\begin{figure}
\centering
  \includegraphics[width=3.5 in]{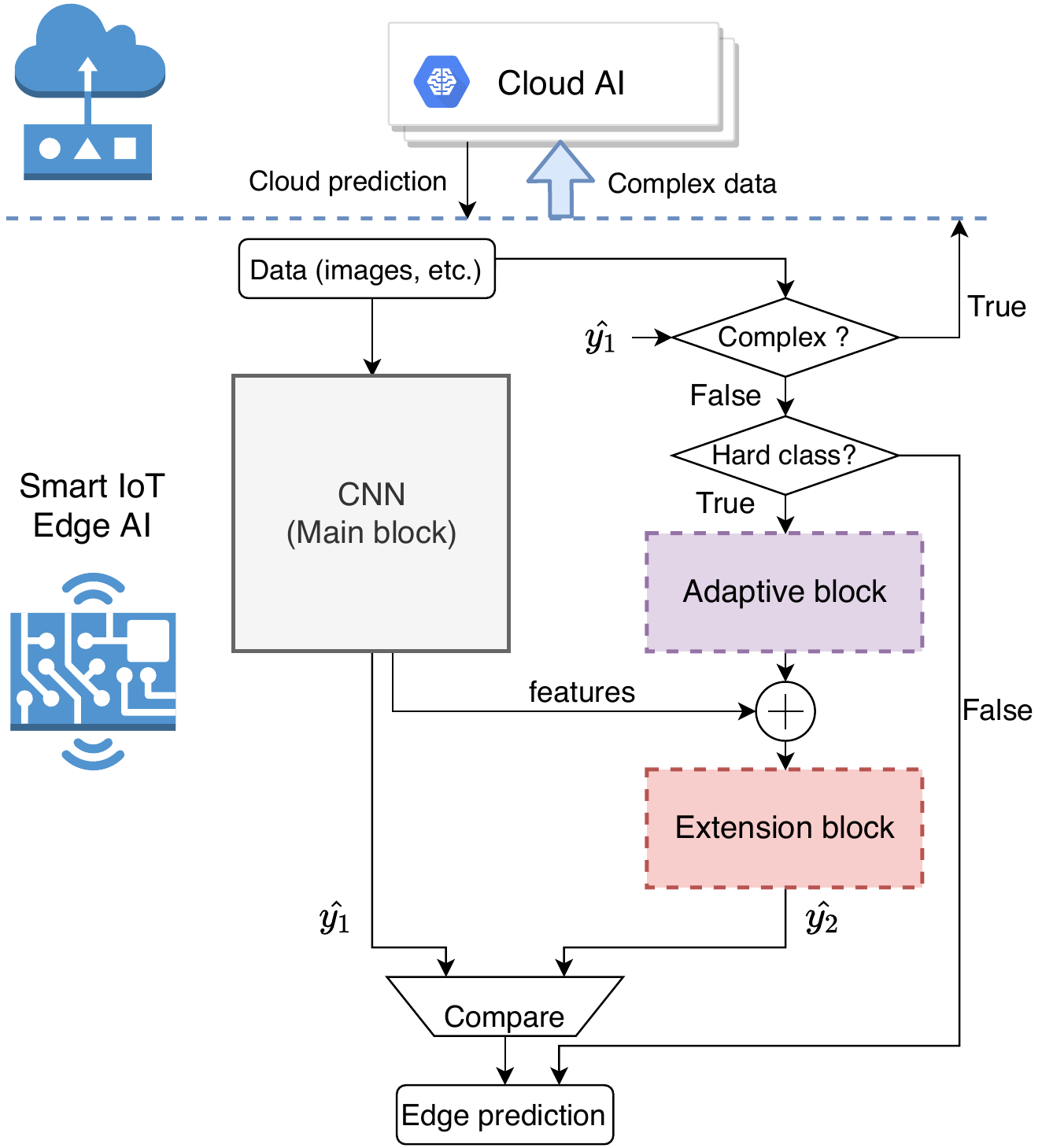}
  \caption{Overview. The distributed system consists of a DNN on the cloud and a MEANet at the edge.}
\label{fig_method}
\end{figure}
%(1. begin with the development/future trend of AI and IoT in smart city/home/industry, and increasing number of data generated by IoT and the requirement to process data intelligently by edge and cloud collaboration. Then explain the high resource/energy requirement of deep neural networks and resource/energy constraint of edge devices, and why data can not be all processed on the cloud. Summarize efficient inference methods.) 
%As the success of Deep Learning attracts more attention in AI, AI has also received considerable attention in the IoT community, where interest in AI has included the investigation of theoretically efficient algorithms and the exploration of numerous applications for smart city/home/industry[ref].

%REORGANIZED: explain why a distributed system is necessary
%TODO: rewrite this paragraph, avoid using the same sentence as in the related work section (DONE)
A distributed AI system with edge-cloud collaboration primarily relies on two pillars: 1) Energy-efficient computing at the edge, 2) Intelligent edge-cloud distribution. Distributed computing consisting of the edge and the cloud has inherent advantages, such as providing system flexibility and scalability, and supporting coordinated central and local decisions \cite{distributed}. Edge computing and cloud computing can collaborate by allowing the edge to activate the cloud as needed. There are two ways of edge-cloud collaboration by conditionally sending the raw data (e.g. images) or processed features to the cloud. The first way allows the edge network and cloud network to be relatively independent. The second approach would partition a deep neural network across the edge and the cloud. The edge network can potentially be simple and suitable for classifying simple data with high accuracy. Various optimization techniques have been proposed to alleviate the computational cost of DNNs for edge computing, such as quantization and branching \cite{Branchy:journals/corr/abs-1709-01686,BinaryNN:journals/corr/CourbariauxB16}. More complex data can be conditionally sent to the cloud based on the confidence derived from the edge network. This can minimize communication and resource usages for edge devices and maximize inference efficiency.

%TODO:(DONE) 1) The research motivation of this paper is not clear. This paper should explain in which scenarios the proposed architecture can be used/ is beneficial. 
%Why training at the edge? Introduce efficient training methods
In addition to distributed inference, since IoT devices continuously collect new data from the environment, distributed training is challenging but beneficial. Although it is possible to upload all data to the cloud for training and then download the updated model, the communication energy and latency would be huge, and the large amount of IoT devices would put significant pressure on the cloud server to respond. Besides, the data privacy is a big concern. Hence, there is a need to explore efficient training algorithms for resource-constrained edge devices. If the model can be adaptively trained with locally collected data, it can better fit the environment, making it possible to outperform the pretrained models. Furthermore, the deployment of pretrained models puts intelligent IoT under the risk of white-box attacks because attackers may access the model parameters when the model is downloaded \cite{attack, IoTML}. Training at the edge modifies model parameters locally, and hence may prevent such attacks.

\begin{figure}[h!]
\centering
  \includegraphics[trim=170 130 170 130,clip, width=2.8 in]{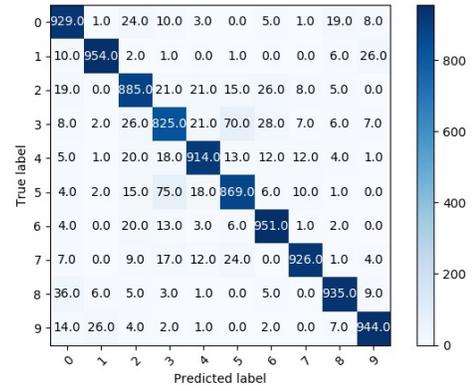}
  \caption{Confusion matrix of running a ResNet32 on the CIFAR-10 dataset. The diagonal of the matrix indicates the number of correct classifications in each class.}
\label{confusion_matrix}
\end{figure}

\begin{figure}
    \centering
    \includegraphics[width=3.0 in]{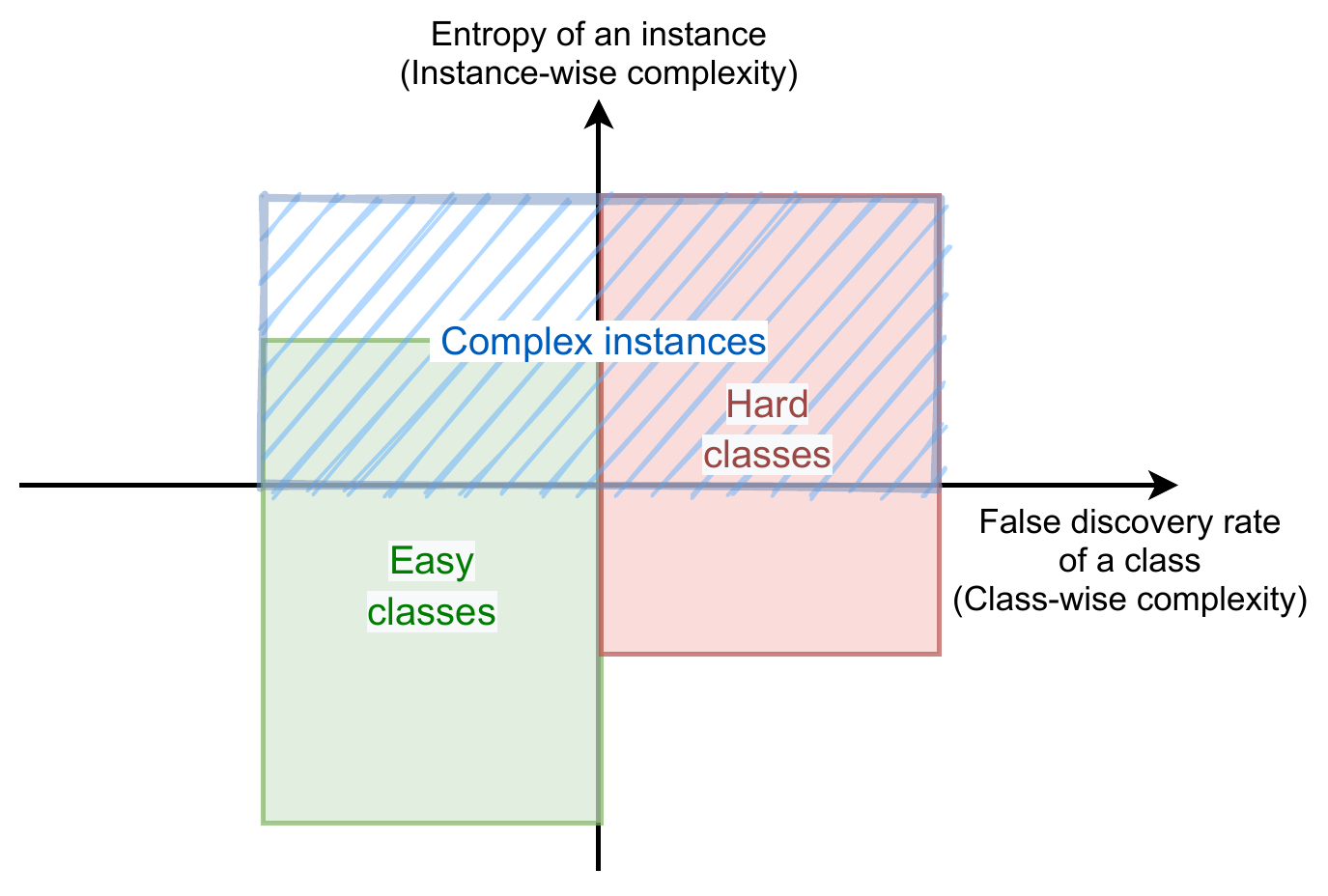}
    \caption{Complexity of instances belonging to easy/hard/complex categories. The false discovery rate (FDR) of every class is evaluated by the main block on the validation set. FDR is the ratio of the number of false positive results to the number of total positive test results ($FDR=1-precision$). The entropy of an instance is evaluated by the main block during inference. Please note that the easy and hard categories have no overlaps, but an instance being either easy or hard may also belong to the complex category.}
    \label{fig:complexity}
\end{figure}

%TODO: (DONE) 3) The paper lacks clear definitions and explanations for instance-wise complexity and class-wise complexity, as well as the easy classes, hard classes and complex instances.
 Due to the resource and energy constraints, it is infeasible to train with a large dataset and do backpropagation in a deep neural network at the edge. To overcome this and realize intelligent edge-cloud distribution, complexity-aware training strategies provide a way to selectively utilize data and parameters. 
% Why complexity-aware strategies
The complexity of data varies widely across instances and classes in real-world datasets, which makes the classification difficulty not uniform \cite{Conditional:journals/corr/PandaSR15}. For example, Fig. \ref{confusion_matrix} shows the confusion matrix of the CIFAR-10 dataset. The precision of some classes is notably lower than others, which reflects their higher \textbf{class-wise complexity}. We classify data into three categories: easy, hard, and complex. Fig.\ref{fig:complexity} shows the definition of these categories depending on the instance-wise and class-wise complexity. The complexity-aware training strategy aims at specifically improving the classification of hard classes and reducing the required amount of training data. On the other side, complexity-aware inference strategies ensure that instances with high \textbf{instance-wise complexity} are sent to the cloud for better accuracy. Early exiting of inference also ensures low latency and energy-efficiency.

%TODO: (DONE) 2) Add design intuition of MEAnet to help readers intuitively understand the design choices and trade-offs considered. There is no detailed description of why MEAnet was designed the way it was (we had description in the third section). Some insight would help inform the reader why it is better than any other design.
%\textcolor{blue}{With the complexity-aware strategies, the next problem is how to design an architecture that reduces the heavy computation and memory usage of deep backpropagation. Our solution is to leverage a pre-trained model and extend it to offer flexibility and adaptivity.}%It might help by fixing the trained weights and weights in the next layers are optimized based on features from the fixed layer. Nevertheless, training sub-models and combining the optimal sub-models does not always produce an optimal deep learning model. The accuracy of such training algorithms has not been as impressive as standard backpropagation.   Our solution is to leverage a pre-trained model and extend it to offer flexibility and adaptivity.}

%% discuss the network architecture and when data (and what kind of data) is sent to the cloud. Then discuss training of such networks.
%meet resource constraints and fulfill accuracy requirements.
In this paper, we design a distributed network architecture that is suitable to apply the proposed complexity-aware training and inference strategies. It consists of an adaptive convolutional neural network (CNN) at the edge and a deep CNN at the cloud. The architecture is shown in Fig. \ref{fig_method}. The edge part is divided into a \textit{main block}, an \textit{extension block} and an \textit{adaptive block}. The architecture leverages a pre-trained main block and extend it by two blocks to offer flexibility and adaptivity. We call this tripartite architecture an ``MEA'' structure and the corresponding network, MEANet. Any deep feed-forward network such as ResNet can be restructured into an MEANet. The main block and the extension block each contain a fully-connected classifier (\textit{exit}). Each exit generates its own prediction $\hat{y}$. 

 As the proposed system combines adaptive MEANet with a deep CNN on the cloud, its training is distributed to the edge and the cloud. The main block of MEANet is trained on the cloud then downloaded to the edge and gives high confidence on inference accuracy of ``easy instances". The classes with lower validation accuracy given by the main block are automatically put in the set of ``hard classes". The main block trained with all classes  ensures the stable performance of the MEANet. The other two blocks of MEANet focus on improving the accuracy of hard classes, and hence they are trained locally with data from hard classes. The main block being fixed, the adaptive block creates a short path for backpropagation and provides features associated with hard classes. With the adaptive block added in the propagation path, the extension block is not completely dependent on the main block, and hence it is able to learn better about hard classes. Using the tripartite architecture and the complexity-aware training strategies, the heavy computation and memory cost of deep backpropagation is reduced.
 
The inference process of MEANet also involves both the edge and the cloud based on the complexity-aware strategies. During inference, each exit controls the activation of the main block, extension block, or the cloud. For example, if an input is detected as an easy class with high probability, it can terminate immediately at the main block. An input detected as hard with high probability would be sent to the other blocks of the edge. If it does not show high probability of belonging to neither easy nor hard classes at the main block, then it is considered complex and is sent to the cloud. After further inference at the cloud, results are sent back to the edge.  Note, the training algorithm and the network architecture lend to automatically determining the classes: easy, hard, and complex. While most instances of easy and hard classes are conditionally inferred at the edge, complex instances are detected at the edge but sent to the cloud for classification. Also note, that the edge device itself comes with conditional inference -- the early exit for easy classes from the main block leads to energy efficient inference at the edge.
%adaptive training edge and assistant inference at the cloud. The memory resources of edge devices are often not enough to train a whole deep neural network. 
%Moreover, the locally collected data may not cover all classes, and hence, a pretrained neural network is difficult to adapt to the edge. Such being the case, we propose to build an edge-cloud AI system that is composed of an efficient CNN at the edge and a deep CNN at the cloud; then train additional neural network blocks at the edge with data belonging to selected classes. Our edge-cloud system enables distributed inference at different stages depending on data complexity. A small portion of data that cannot be handled by the edge (due to resource or energy constraint) is sent to the cloud to improve the accuracy. Finally, we evaluate different ways of edge-cloud collaboration in terms of energy, latency, and data privacy.
%emphasis insufficient training data/resources at edge % saving training efforts

The main contributions of this paper are
\begin{itemize}
\item We propose a novel architecture, MEANet, that combines the pretrained main block with locally trained extension and adaptive blocks. MEANet at the edge, together with a DNN at the cloud, form a distributed system to smartly manage and process data.
\item We propose complexity-aware methodologies for distributed training and inference. The distributed system considers both class-wise complexity and instance-wise complexity of data. The proposed methodologies are applicable to a wide range of deep learning models.
   \item We discuss and evaluate different ways of edge-cloud collaboration and their advantages. We compare edge-cloud distributed approaches with edge computing and cloud computing in terms of accuracy and energy consumption.
\end{itemize}

The rest of this paper is organized as follows. Section II summarizes the related works. Section III describes the details of the proposed model along with the training and inference algorithms. Then we discuss different ways of edge-cloud collaboration. Section III shows the implementation details and analyzes the experimental results. Conclusions are drawn in Section IV.

%talk about the resource limitation of edge devices & resource requirement of deep learning models, then the need for an edge-cloud system

%TODO: Reviewer's comments 1) The research motivation of this paper is not clear. This paper should explain in which scenarios the proposed architecture can be used/ is beneficial. 2) Add design intuition of MEAnet to help readers intuitively understand the design choices and trade-offs considered. There is no detailed description of why MEAnet was designed the way it was. Some insight would help inform the reader why it is better than any other design. 3) The paper lacks clear definitions and explanations for instance-wise complexity and class-wise complexity, as well as the easy classes, hard classes and complex instances. 4)no detailed comparisons of the timing as well as energy comparisons. Include additional analyses on key metrics including computation cost for online model training 5) Elaborate the experiment methodology description, to improve reproducibility 6) Have a standalone related work section that explicitly discusses this work in the context of other important related work. A list of papers we suggest authors consider is attached at the end.

\section{Related Work }
\subsection{DL Algorithms and Architectures for Edge Computing}
To make deep learning available for edge/mobile computing, efficient algorithms and architectures are proposed, including quantization, pruning, branching, MobileNets, and neural architecture search. Quantization methods involve reducing the precision of weights and activations \cite{Quantization:journals/corr/abs-1712-05877, BinaryNN:journals/corr/CourbariauxB16, XNOR:journals/corr/RastegariORF16, zhou2016dorefa, chakraborty2020constructing, choi2018pact}. Pruning is another technique \cite{han2015deep, li2016pruning, garg2019low} of reducing the number of parameters and computational complexity. Branching creates dynamic inference paths to conditionally activate layers according to the difficulty or property of the input, in order to improve efficiency or accuracy. Multi-branch neural networks, such as BranchyNet \cite{Branchy:journals/corr/abs-1709-01686}, use different training, early-exiting, and dynamic routing strategies \cite{MVNet,dynamic,BlockDrop, EPNet,ImprovedAdaptive}.
MobileNets, which use depthwise separable convolutions, allow shrinking the model by global hyper-parameters to match the resource restrictions for specific applications \cite{Mobilenet:journals/corr/HowardZCKWWAA17}\cite{Mobilenetv2:journals/corr/abs-1801-04381}. 
Neural architecture search first finds the best network architecture in a search space\cite{NAS:journals/corr/ZophVSL17}, then deploy it to the edge.

%Training
Several existing training algorithms apply the idea of divide-and-conquer to enable efficient training. Incremental learning and continual learning use a continuous learning process to accommodate previously unseen data and tasks \cite{increm_2017_CVPR,incremental,continual1,continual2}. To alleviate catastrophic forgetting in continual learning, different studies can be partitioned into architectural, functional, and structural approaches. For example, an episodic memory can be used to store a subset of observed data\cite{episodic_memory}. Supervised pretraining aims to modify deep learning models into simpler versions that are easier to train \cite{scaling}. Before powerful GPUs became available, this was one of first attempts to train deep networks. Nevertheless, combining the optimal sub-models does not always produce an optimal deep learning model. Similarly, Bloctrain has been proposed to freeze trained blocks while gradually adding new blocks for training spiking neural networks \cite{bloctrain}. 
%For adaptive architectures with multiple intermediate classifiers, \cite{ImprovedAdaptive} presents three training techniques to resolve the conflict between classifiers and enhance their collaboration.

\subsection{Distributed Deep Learning}
%distributed ML(edge/system)
Distributed DL includes distributed inference and training for a multi-node system or a edge-cloud system.
 The main idea of distributed DL for the edge is to partition a deep neural network or coordinate multiple shallow neural networks. Layer-based partitioning of DNNs is a natural approach that applies branching, but distributing multiple branches at different nodes would result in transmitting a large amount of intermediate data. Another way to distribute a DNN is to divide each convolutional layer into independently distributable tasks, thus enabling parallel processing \cite{deepthings}. TeamNet provides a mixture-of-experts approach which trains multiple expert models with data selected by a gate network\cite{TeamNet}. %However, it requires broadcasting data to all experts which brings a heavy communication overhead. 
 Large-scale distributed DL systems with multiple nodes can achieve high performance by parallelized optimization algorithms\cite{HPDL, PSGD,parameterserver, largescale,4minsDBLP:journals/corr/abs-1807-11205}. %Since multiple nodes are included, the distributed DL systems apply either tree-based parameter server \cite{parameterserver, largescale} or ring-based all-reduce communication mechanism \cite{4minsDBLP:journals/corr/abs-1807-11205}.

%distributed NNs across edge and cloud
%The communication mechanism in an edge-cloud distributed system is simpler than that in a multi-node system.  Depending on whether the cloud network is independent or connected with the last layer of the edge network, the edge sends either raw data or processed features to the cloud.
Besides improving computing efficiency and speed, the other aspect of distributed DL is edge-cloud collaboration. In \cite{distributed} and \cite{mobilecloud}, an intermediate layer with lightweight features is selected as a partition point between the edge and the cloud to save communication cost. In \cite{Neurosurgeon}, a scheduler called Neurosurgeon is proposed to automatically partition DNN computation
between mobile devices and datacenters at the granularity
of layers. SPINN is a distributed inference system that co-optimizes the early-exit policy and DNN partition at runtime, in order to adapt to dynamic conditions \cite{SPINN}. Distributed DL can be combined with quantization techniques to build hybrid networks, which consist of low precision layers at the edge and full precision layers on the cloud \cite{ours}. Training can also be distributed across the edge and cloud. For example, gradient-descent based distributed learning algorithms involve local update steps at edge nodes and global aggregation steps performed by the cloud\cite{DistributedML}. %Each edge node performs gradient descent to adjust the local model parameter; then the cloud periodically receives parameters obtained at different edge nodes and sends back the average parameters.

\section{Complexity-aware adaptive neural networks}
%summary
\begin{figure*}
\centering
  \includegraphics[width=5.6 in]{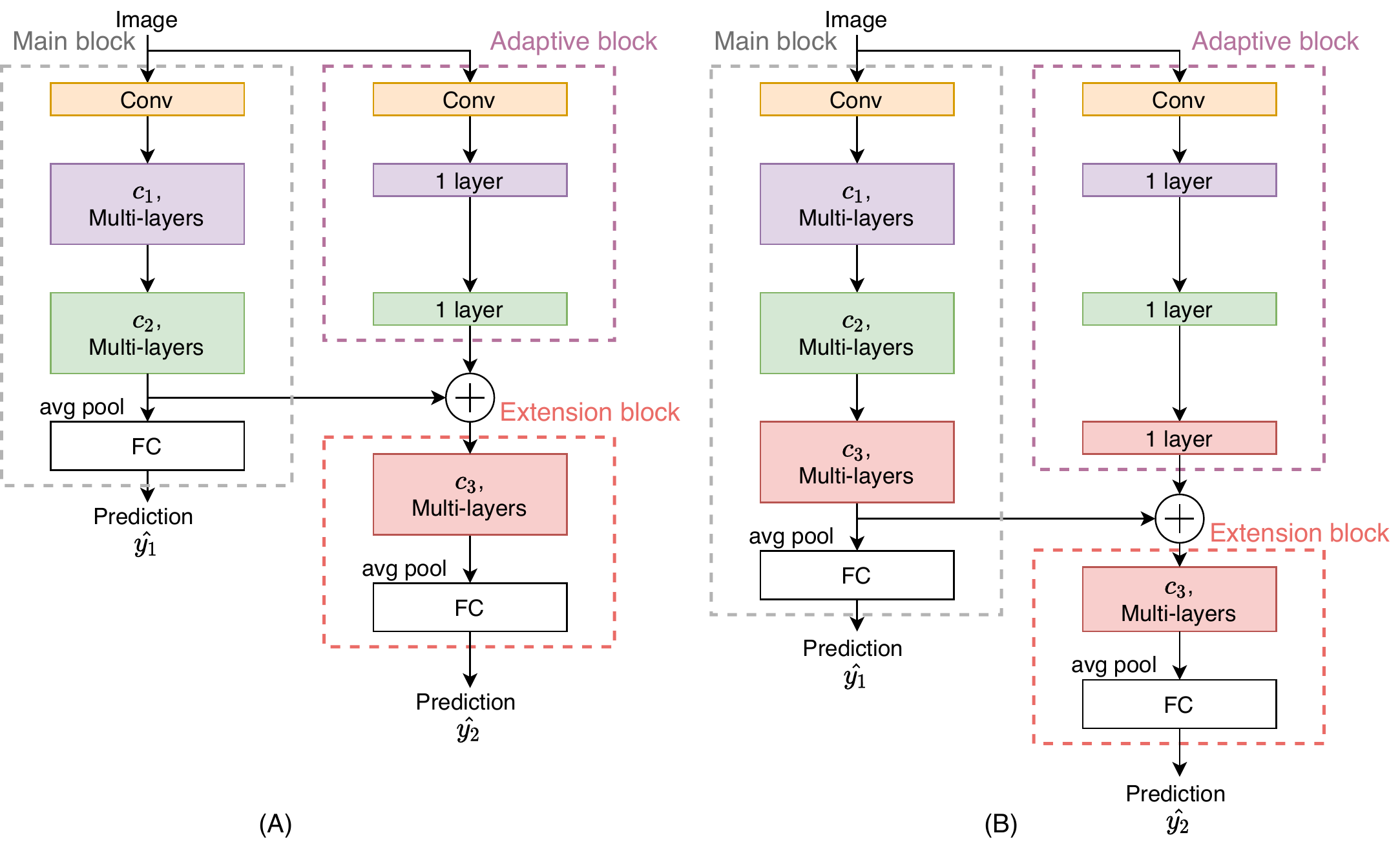}
  \caption{Examples of MEANets. $c_1$, $c_2$, $c_3$ are the numbers of output channels of convolutional layers. Having the same color means having the same number of output channels and kernel size.}
\label{fig_example}
\end{figure*}
%In this section, we first describe the architecture of the proposed neural network and the training and inference techniques, then we discuss different ways of edge-cloud collaboration and compare them.
\subsection{Training}
The motivation for training at the edge is to improve the adaptivity of the edge device and to enhance its perception on data. However, we must consider the resource and energy constraints. Since backpropagation in the training process is computationally expensive, training a deep neural network at the edge is challenging. Also, it is impracticable to store the entire dataset at the edge or require the IoT devices to collect data of all classes. Furthermore, while having multiple exits enables complexity-aware learning, the optimization of these exits requires a special training algorithm. Hence, it is essential to design an adaptive architecture and a suitable training algorithm for the edge.

First, let us describe the architecture of the MEANet, which is shown in the bottom part of Fig. \ref{fig_method}. The main block contains the majority of convolutional layers of the edge neural network. Because of the large number of parameters, it is preferred to train network in the cloud, then deploy to the edge. Having obtained a pretrained main block, the extension block is added to increase the depth of the network. It produces classification results $\hat{y_2}$ in addition to $\hat{y_1}$ from the main block.  Although the main block is able to perform classification, it yields lower precision in hard classes. The extension block brings more layers to help making decision in hard classes. With the assistance of the extension block, MEANet has stronger learning ability to deal with complex data.

The purpose of using the adaptive block is to create a short path for backpropagation. The problem with training only the extension block is that it is strongly affected by the main block. Although we optimize the extension block for the loss at the local exit, the inputs to the block are features generated by the fixed main block. Thus, it is likely to perform the same misclassifications as the main block. It is important to make sure that the extension block is trained based on the raw inputs, which are independent of parameters at the main block. Hence, we use an adaptive block to connect the extension block with the raw inputs. The outputs of the adaptive block have the same size as those of the main block. Then the sum or concatenation of them are used as the inputs to the extension block. If concatenation is chosen, the first layer in the extension block would have more input channels. Compared to the main block, the adaptive block is designed to be much shallower to limit the corresponding computational and memory overhead. %At the edge training stage, the extension block and the adaptive block are trained based on only samples in hard classes instead of the whole dataset, therefore the training time and energy are reduced. 

Two examples of the MEANet are shown in Fig. \ref{fig_example}. The adaptive block consists of convolutional layers similar to those in the main block. The number of out channels of each layer in the adaptive block matches with the corresponding layer in the main block. Therefore, their outputs have the same size. In other words, the adaptive block is a light-weight version of the main block. 
In model A, a typical CNN such as ResNet is divided into two parts. The first part is used as the main block and the second part including the fully connected layer is used as the extension block. An extra fully connected layer (exit) is created for the main block. In this case, the main block has fewer layers, so it is possible to also train it at the edge. In model B, a complete CNN is used as the main block, then the adaptive block and extension block are added. Model B has a deeper main block than the model A. Consequently, it generally achieves better accuracy, but its main block must be trained at the cloud before downloading to the edge. Note, the additional blocks can always be trained at the edge.

%summarize the training algorithm
%emphasize that in the forward pass, both branches are active, while in the backward pass, only the adaptive block is active
\begin{algorithm}[h]
\SetAlgoLined
\KwIn{Instances $X$ and labels $Y$}
\KwOut{Trained neural networks}
 1. Train a deep CNN as the cloud AI. Also train the main block of the edge AI at the cloud with the whole dataset.\\
 2. Run evaluation on the validation set to determine hard classes $C_{hard}$.\\
 3. Create a dictionary ClassDict to map labels of all classes to new labels of hard classes.\\
  label = 0\;
    \For{c in $C$}{ 
        \If{$c\in C_{hard}$}{
            ClassDict[c]= label\;
            label+=1\;
        }
   }
 4. Download the main block and ClassDict to the edge.\\
 5. Select training data $X_{hard}$ and generate new labels $Y_{hard}$.\\
  \For{instance i in $X$}{ 
        \If{$Y$[i]$\in C_{hard}$}{
            Index.append(i)\;
            $Y$[i]= ClassDict[$Y$[i]]\;
        }
   }
  $X_{hard}$ = X[Index];  $Y_{hard}$ = Y[Index]\;

 6. Add the adaptive block and the extension block to the neural network. Fix the main block.\\
 7. Forward propagation.\\
  $\hat{y_1}, F =$ Main-block($X_{hard}$)\;
  $f_2=$ Adaptive-block($X_{hard}$)\;
  $\hat{y_2}$ $=$ Extension-block($F$,$f_2$)\;
 8. Backward propagation.\\
  Loss = CrossEntropyLoss($\hat{y_2}$, $Y_{hard}$)\;
  Loss.backward()\;
    
 \caption{Distributed training of the edge neural network}
\end{algorithm}
Having described the adaptive architecture, we then propose a training algorithm for the MEANet. To save training efforts at the edge, the training algorithm reduces training data and uses blockwise optimization. There are three different methods to train multiple exits in a neural network: joint optimization, separate optimization, and blockwise optimization. Joint optimization optimizes the weighted sum of losses at all exits\cite{Branchy:journals/corr/abs-1709-01686}. Separate optimization trains all convolutional layers based on the loss at the final exit, then freezes them and trains the other exits. Blockwise optimization divides the neural network into blocks. Each block contains several convolutional layers and an exit. When a block is trained, all blocks ahead of it are frozen, leading to improvement in training time and training energy. 
%compare them
 In our training algorithm, we use blockwise optimization for the following reasons. Although using joint optimization achieves the best accuracy, it is not suitable for training at the edge. The edge can hardly afford to train all parameters of a deep network at the same time. In order to address the challenge of training with resource constraints, blockwise optimization is preferred. It only requires to store gradients of the non-fixed parts for backpropagation. Consequently, it largely reduces the memory requirement and energy consumption. Moreover, by fixing the trained main block, it avoids unnecessary training efforts and secures the learned properties.

The detailed steps of the distributed training algorithm are described in Algorithm 1. First, the main block is trained at the cloud with the entire dataset. Next, it is run on the validation dataset to produce class-level statistics. Then we rank the classes according to their precision in increasing order and define first $N_{hard}$ classes as hard classes, where $N_{hard}$ is a user-defined parameter. Other classes are defined as easy classes. Because the labels of hard classes are not likely to be consecutive in the set of all classes $C$, we generate a new set of labels exclusively for hard classes and call it $C_{hard}$. Then the set of easy classes is given by $C-C_{hard}$. A dictionary is created to record the mapping of $C$ and $C_{hard}$. After that, we download the main block to the edge and train the extension and adaptive blocks. Since the main block has been trained, it is fixed while training the other blocks. The extension and adaptive blocks aim to deal with complex cases, therefore, it is unnecessary to train them on the entire dataset. Instead, we only use instances in hard classes. The labels of instances are checked and only those belonging to $C_{hard}$ are used for training. As a result, the number of classes is reduced and the size of training dataset that is usually proportional to the number of classes becomes smaller. More importantly, the weights of the adaptive block provide features specifically for hard classes. By learning from these features, the extension block is able to correct the decision boundaries suggested by features from the main block. The remaining question is how to detect instances of hard classes and manipulate the activation of these blocks during inference. We will describe this in the next subsection. 
 %As the motivation for adding the extension block is to have a deeper network with stronger learning ability, a class with lower precision should have higher priority because this implies higher class-wise complexity.

Please note that in our experiments, we use the hard-class samples from the training dataset to train the extension and the adaptive blocks. This simulates the case that data collected from the environment have the same distribution as those in the dataset. In the real environment, the edge can collect new samples that have a different distribution. To avoid overfitting and catastrophic forgetting on the new samples, we suggest using both the new samples and samples from the dataset for training.

% Add more layers to deal with classes with lower accuracy. Fix the trained neural network and only train the additional layers so that the cost of backpropagation can be saved. 

\subsection{Inference}
\begin{figure}
\centering
  \includegraphics[width=3.4 in]{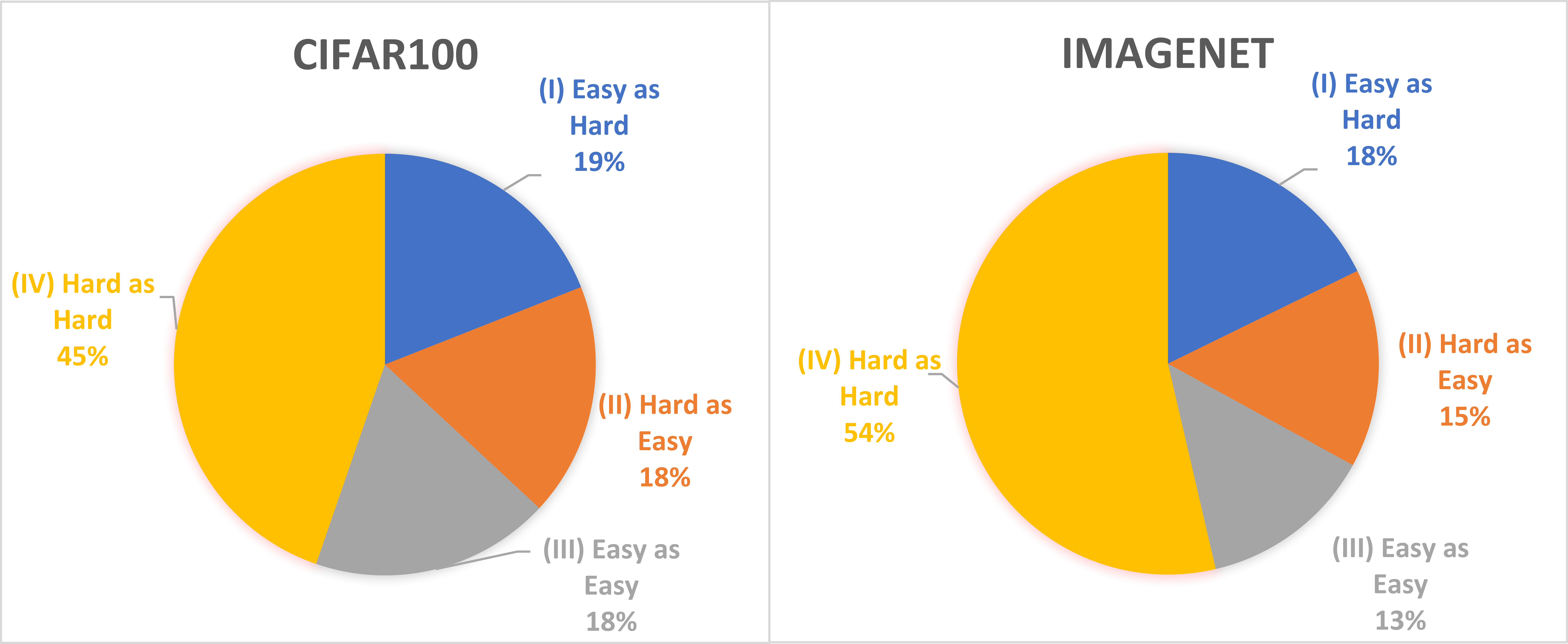}
  \caption{The proportion of four types of errors: (I) mistaking a sample of a easy class as hard;  (II) mistaking hard as easy;  (III) mistaking a sample of a easy class as another easy class;  (IV) mistaking a sample of a hard class as another hard class. Here we assume half of classes are hard. }
\label{fig_mistake}
\end{figure}

During inference, whether an instance belongs to one of the hard classes or easy ones is decided based on the prediction by the main block. Although it is optional to train a binary classifier as a detector, we find that using the outputs of the main block to detect easy/hard classes is the simplest and the most effective way. %TODO: mention other options and explain why we choose this
The complexity-aware partition of classes ensures accurate detection since the average precision on easy classes is higher than that on all classes. We denote the output vector of the main block by $\hat{y_1}$, whose length equals the number of classes. Then the softmax function is applied to get the probability $p_1$ of belonging to each class. The predicted class is the index of the maximal element in $p_1$. If the predicted class is one of the easy classes, the corresponding instance is classified as belonging to easy classes and is allowed to exit at the main block; otherwise, the other two blocks are activated. The detection of easy/hard classes can be summarized as
  \begin{equation*}
  p_1=softmax(\hat{y_1})
    \end{equation*}
 \begin{equation*}
    IsHard(\hat{y_1})=
    \begin{cases}
      True, & \text{if}\ argmax (p_1)\in C_{hard} \\% & \text{and}\  \max(p_1)> threshold1 \\
      False, & \text{otherwise}
    \end{cases}
  \end{equation*}
The detection of easy class vs. hard class is not 100\% correct, however, the misinformation is not fatal due to the following reasons. Fig. \ref{fig_mistake} depicts the proportions of four types of errors when evaluating the main block on CIFAR100 and ImageNet. The four types are (I) mistaking a sample of a easy class as hard;  (II) mistaking hard as easy;  (III) mistaking a sample of a easy class as another easy class;  (IV) mistaking a sample of a hard class as another hard class. Around half of the errors by the main block belong to type IV. The extension block trained on hard data can improve type IV errors. As long as the extension block makes considerable improvement on this type, the overall accuracy will increase accordingly. If the IsHard function is modified to consider more samples as hard, some type II errors will be turned into type IV, which may take more advantage of the extension block. However, adjusting the detection function would also cause sending instances of easy classes to the extension block by mistake. In such cases, the extension block is not able to classify these instances correctly since it is trained with only hard classes. Therefore, modifying the IsHard function can not guarantee an overall improvement. Nevertheless, if the main block confuses easy classes with hard ones, then its prediction is wrong and the prediction by the extension block could not be worse. Hence, although the imperfect detection limits the improvement brought by the extension block, it is acceptable.

%TODO: abandon the threshold
%The threshold for confidence is dependent on the number of classes. For example, with 1000 classes (say), the average confidence of predictions on the dataset would be smaller than that with 10 classes. Choosing an appropriate threshold is important because if it is too high, the number of false negatives will increase which causes mis-classification in the extension blocks. On the other hand, if it is too low, the number of false positives will increase and the improved decision boundaries of the extension block will not be fully used. 
After the extension block provides new predictions, we compare predictions from the two exits at the main block and the extension block and select the one with higher confidence. The softmax score of the predicted class, which is also the maximal softmax score among all classes, indicates the confidence on the prediction. This is to ensure the final prediction improves compared to the prediction by the main block. Furthermore, the wrongly detected instances belonging to complex instances can be handled by the cloud. We will describe the details in the next subsection.

%Training with a reduced number of classes can save training energy and make the classification problem easier such that the extension block can classify instances of hard classes more precisely. 

\begin{table*}[t]
    \centering
    \begin{tabular}{|c|c|c|c|c|}
 \hline

  & Edge computation  & Cloud computation & Communication& Data Privacy \\
 \hline
Edge  & $N\cdot x$ & / & / & ++++\\
 \hline
Cloud  & / & $N\cdot x_{cl}$ & $N\cdot x_{cu}$& +\\
 \hline
Edge-cloud (sending raw data) & $N\cdot x$ & $\beta \cdot N\cdot x_{cl}$ & $\beta \cdot N\cdot x_{cu}$ & ++\\ 
 \hline
Edge-cloud (sending features) & $N\cdot(qx)$ & $\beta \cdot N\cdot(1-q)x_{cl}$ & $\beta \cdot N\cdot x_{cu}'$ & +++\\ 
 \hline
    \end{tabular}
    \\[8pt]
    \caption{Cost estimation of inference using edge, cloud and edge-cloud collaboration. $N$ is the total  number  of  data  instances. $x$ is the energy consumption or the latency to process an instance at the edge. $\beta$ is the percentage of data sent to the cloud. $x_{cl}$ is the cost of cloud computation. $x_{cu},x_{cu}'$ are the cost of communication. $q$ is the proportion of layers distributed at the edge.}
    \label{edge-cloud-compare}
\end{table*}
%\subsection{Fast Inference}
%Choose efficient neural networks for edge AI: conditionally deep hybrid neural networks and etc.\\
%not only for energy efficiency, it enables edge to analyze data locally (even if the cloud is not responding, it can use rough results from its own network). 

\subsection{Distributed inference by edge and cloud}
%explain how cloud and extension collaborate to handle both instance and class wise complexity
The edge model can either infer data independently, which ensures data security, or ask a deeper CNN at the cloud to deal with complex instances. The design methodology for MEANet improves the accuracy of hard classes, however, there are complex instances in both easy and hard classes. Hence, both the extension block and the cloud are needed to deal with class-wise complexity and instance-wise complexity, respectively. Making edge and cloud collaborate for inference can improve the tradeoff between energy and accuracy by exploiting edge resources and occasionally activating cloud resources. Note, it is important to utilize edge resources to analyze data locally as much as possible for faster inference. Although the deeper network at the cloud is more accurate, a large portion of instances might have been predicted correctly at the edge. Therefore, sending them to the cloud is a waste of energy and time. Furthermore, having the option of sending to the cloud solves the misdetection of easy/hard classes. Since the misdetected instances do not belong to any classes at the extension block, they should be directly sent to the cloud.

The entropies of predictions at the main block are used to decide which instances to send to the cloud, similar to that in \cite{distributed,ours}. The entropy of a prediction indicates the instance complexity and the confidence of the classifier. At the main block, the entropy values of correct ones show an exponential distribution peaking at zero, while those of wrong predictions show a normal distribution whose mean is larger than one.  Instances with entropy higher than a threshold are sent to the cloud. By evaluating the entropy values of the validation set, the range of the threshold can be determined as $(\mu_c,\mu_w)$. Here $\mu_c$ is the mean of entropy of correct predictions and $\mu_w$ is that of wrong predictions. Then the user can select a threshold in the range based on the system requirements. The higher the threshold is, fewer instances are sent to the cloud and the inference is faster and more efficient.
%At the main block, the entropy values of wrong predictions show a normal distribution while those of correct ones show an exponential distribution. By evaluating the entropy values of the validation set, statistical models are built to approximate the entropy values of the test set. Then the threshold is determined by choosing the number that maximizes the cumulative distribution of correct predictions minus that of wrong predictions. The objective function is
 %\begin{equation*}
 %    N_{c}* \int_{0}^{thre} f_{exp}(x,m_c) dx - N_{w} *\int_{0}^{thre}f_{normal}(x,m_w,\sigma_w)dx
 %\end{equation*}
 %where $f_{exp}$ and $f_{normal}$ is the probability density function for the exponential and normal distribution. $N_c$ and $N_w$ are the number of correct/wrong predictions in the validation set. $m_c, m_w, \sigma_w$ are the mean and standard deviation of entropy with regard to correct/wrong predictions.}
 %Thus it exploits the capability of the edge. Note, only instances predicted as belonging to hard classes are considered for sending to the cloud. An instance sent to the cloud is either an instance of a hard class or a wrongly detected instance of an easy class. Considering both class-wise complexity and instance-wise complexity reduces the frequency of the cloud being activated and saves communication cost. 

There are two ways of edge-cloud distributed inference. Either raw  data or processed features are sent to the cloud depending on the architecture of the cloud network. In many distributed learning work, features are sent because the layers at the edge and cloud are components of a partitioned deep neural network\cite{distributed}. When the component at the cloud receives output features from the edge as inputs, it continues processing and sends back results. This approach seems natural, but sending features forces the data processing at the cloud to be dependent on the data processing results from the edge. On the other hand, if raw data are given, the cloud AI can be independent of the edge AI. Hence, a more accurate network can be used at the cloud to improve the accuracy and the edge-cloud system is more flexible. For this reason, we use a ResNet101 as the cloud network and send raw data to the cloud.

 Algorithm 2 shows the overall process of inference in the edge-cloud distributed system. In summary, all instances are first processed by the main block. Then instances with high entropy (low confidence) are sent to the cloud. Other instances are allowed to exit at the main block if they are classified as belonging to easy classes; otherwise, they are sent to the extension block.
 \begin{algorithm}[h]
\SetAlgoLined
\KwIn{A batch of images $X$ from the dataset}
\KwOut{Classification results $\hat{Y}$, features $F$}

 \For{a batch of instances}{
    $\hat{y_1}, F =$ Main-block($X$)\;
    \For{instance i in $X$}{ 
        
        \If{Cloud is available and Entropy($\hat{y_1}$[i]) $>$ threshold }{
                Send instance I[i] or feature $F$[i] to the cloud\;
                Receive the classification result\;
                $\hat{Y}$[i] $= \hat{y_{cloud}}$[i]\;
                continue\;
        }
        \eIf{IsHard($\hat{y_1}$[i])}{
            $f_2=$ Adaptive-block($X$[i])\;
            $\hat{y_2}$[i] $=$ Extension-block($F$[i]+$f_2$)\;
            \eIf{Confidence($\hat{y_1}$[i]) $>$ Confidence($\hat{y_2}$[i])}{
                $\hat{Y}$[i] $= \hat{y_1}$[i]\;
            }{
                $\hat{Y}$[i] $= \hat{y_2}$[i]\;
            }
        }{
                $\hat{Y}$[i] $= \hat{y_1}$[i]\;
                continue\;
            }
    }
 }
 \caption{Inference in an edge-cloud AI system}
\end{algorithm}

\subsection{Cost estimation for inference}

%Also, if the processed features are sent to the cloud, the cloud does not need to perform similar executions as the edge. However, the computation energy at the cloud may not be of big concern. Note, the two approaches have their advantages and disadvantages and they should both be considered for IoT applications.

In Table. 1, the cost of edge-cloud distributed inference is compared with that of edge inference or cloud inference. The cost can be energy consumption or latency. Using this table, one can estimate the energy or latency of inference by inserting values of variables. Total cost is broken down into the computation cost of the edge, the computation cost of the cloud, and the communication cost. Assume that the total number of data instances is $N$. The computation cost to process an instance at the edge and that at the cloud are $x$ and $x_{cl}$. Then the total cost of edge computation is $N\cdot x$. The communication cost can be either $x_{cu}$ or $x_{cu}'$ corresponding to sending raw data or features respectively. Whether sending raw data or features consumes more communication energy depends on the dataset. If the size of data in a dataset is small (e.g. CIFAR datasets), the size of features is usually larger than that of raw data. In the ImageNet dataset, the size of raw data might be larger. With early exiting allowed by edge-cloud collaboration, the percentage of data instances sent to the cloud is reduced to $\beta$ in the range of $[0,1]$. In the case of distributing layers and sending features to the cloud, the proportion $q$ of layers distributed at the edge needs to be considered (typically $q\in [1/3, 2/3]$). In the next section, we will estimate energy consumption using this table.

%Using this table, one can estimate the energy of different edge-cloud AI systems by inserting proper values of variables. Also, if the definition of $x$ is changed to latency, this table can be used to estimate the latency in an edge-cloud AI system. Such considerations are evaluated in the next section. For the proposed edge-cloud system, we chose sending raw data because it allows the usage of an independent deep network at the cloud.

%Utilize edge prediction and cloud prediction to ``vote" on the final prediction.

\section{Experimental results}

\subsection{Setup}
To validate the effectiveness of our methods, we conduct several sets of experiments with different CNN architectures and datasets. We use ResNet\cite{he2016deep} and MobileNetV2\cite{Mobilenetv2:journals/corr/abs-1801-04381} as the basic architectures and extend them as described in section II. As image classification being one of the most popular computer vision tasks, we test the models on two image classification datasets, CIFAR-100\cite{cifar} and ImageNet\cite{ImageNet}. 

We build a simulation environment using PyTorch to simulate edge-cloud distributed AI systems. The statistics including the overall accuracy, the accuracy of hard classes, and the percentage of instances exiting at each exit are recorded. Then the energy consumption for communication and computation is estimated.

The architectures of the models are illustrated in Fig.\ref{fig_example}. Because CIFAR dataset contains much fewer classes and images, the convolutional layers of ResNet used for CIFAR100 have 16, 32, and 64 channels respectively from the first group of layers to the last one ($c_1=16, c_2=32, c_3=64$). Every group contains the same number of  convolutional layers. The small number of channels used for CIFAR100 makes it suitable for smart IoT edge. As described in section II, model A uses the a part of the original ResNet as the extension block, while model B keeps the complete ResNet as the main block and add new layers to be the extension block. After modification, we add the letter A or B in the name of the ResNet to show which type is used. 

The ResNet used for ImageNet has four different groups of convolutional layers -- 64, 128, 256, and 512 channels respectively. A pretrained ResNet18 is downloaded from PyTorch official website as the main block, then the adaptive and extension blocks are added and trained to change it into model B. %MobileNet
We also apply the proposed method (model B) to MobileNetV2. Similar to what we do to ResNet, MobilenetV2 pretrained on the ImageNet dataset is used as the main block.  The other blocks are subsequently added. To limit the computational overhead, the extension block for model B is designed to have four residual blocks.

%Training epoch and learning rate, validation set
For both datasets, 10\% of the training data are used as the validation set to determine hard classes. The proportion of the training set used to train the extension and adaptive blocks depends on the number of hard classes. The default number of hard classes $N_{hard}$ is half of the total number of classes, and consequently half of the training set is used. After training the main block or downloading the pretrained model, we add the adaptive and extension blocks to the model. In every training epoch, the model is set to train mode, while the layers in the main block are set to evaluation mode, and the corresponding parameters are set not to require gradients. Please refer to Section III for the detailed training steps. For CIFAR-100 experiments, the models are trained with initial learning rate equaling 0.1, then multiplied with 0.1 at epoch 60, 120, and 160. For ImageNet experiments, the models are trained with initial learning rate equaling 0.01, then multiplied with 0.1 at epoch 30 and 100.

\subsection{Results}
\begin{table}[h!]
    \centering
%memory usage; latency; energy analysis
\begin{tabular}{ |c|c|c||c|c| } 
\hline
\multirow{2}{*}{Dataset, model}& \multicolumn{2}{c|}{Train } & \multicolumn{2}{|c|}{Test } \\
 \cline{2-5}
 & main & MEANet & main & MEANet \\
  \hline
CIFAR-100, ResNet32 A &   69.84 & 97.22  & 50.70 &  59.76  \\ %updated
  \hline
CIFAR-100, ResNet32 B & 77.95 & 96.07 & 59.36 &  63.66   \\ 
  \hline
ImageNet, MobileNetV2 B & 65.25 & 70.53 & 61.26 &  65.26 \\%TODO:update train 71.27
  \hline
ImageNet, ResNet18 B & 61.01 & 72.33  & 59.98 &  64.95  \\  %TODO:update train 74.18
 \hline
 \end{tabular}
\\[8pt]
    \caption{Accuracy of hard classes (\%)}
    \label{hard_compare}
%\\[8pt]
\end{table}

\subsubsection{Results on hard classes (edge)}
Because the learning process at the edge only uses training data belonging to hard classes, it is reasonable to first look at the training and testing accuracy of hard classes before and after adding the extension and adaptive blocks. In the tables, we call the results corresponding to the main block as ``main" and those corresponding to MEANet as ``MEANet". The results in Table. \ref{hard_compare} are evaluated using only half of instances in the datasets which belong to hard classes. This simulates the case that the edge can only get data in these classes from the environment. Under this circumstance, the extension and adaptive blocks are always activated. The training accuracy of hard classes is significantly increased after adding the extension and adaptive blocks. Compared to the original model, the test accuracy of ours increases by 4-9\% for CIFAR-100 and 4-5\% for ImageNet, dependent on different architectures. Because the main block in model A has fewer layers compared to that in model B, the extension block is able to provide larger accuracy improvement than model B. The results show that our proposed model has a stronger ability to distinguish between hard classes.

\subsubsection{Results on all classes (edge)}
\begin{table}[h!]
    \centering
\begin{tabular}{ |c|c|c||c| } 
\hline
Dataset, model & main & MEANet & easy/hard detection\\
  \hline
CIFAR-100, ResNet32 A & 61.70  & 63.51   & 83.47  \\ 
  \hline
CIFAR-100, ResNet32 B & 67.55  & 67.80   & 87.88 \\ 
  \hline
ImageNet, MobileNetV2 B & 71.87  &   73.19  & 90.64\\ %71.98
  \hline
ImageNet, ResNet18 B & 69.75 & 71.62 & 88.49 \\ %70.17
 \hline
 \end{tabular}
\\[8pt]
    \caption{Test accuracy(\%) of all classes}
    \label{all_compare}
%\\[8pt]
\end{table}
Table. \ref{all_compare} lists the accuracy of the MEANet when the entire test dataset is evaluated. The overall accuracy on ImageNet increases by nearly 2\% after adaptively training the MEANet. For the CIFAR dataset, Model A provides a larger improvement than model B because its main block is shallower and underfits the dataset. There are two reasons for the improvement being insignificant compared to the results on hard classes. First, because our model focuses on increasing the accuracy of hard classes, the increase is evened out when all classes are presented. Second, the misdetection of easy/hard classes by the main block weakens the strength of the extension block. To compensate for this, edge-cloud collaboration can be used. In conclusion, the advantage of our model is maximized if the main block underfits the dataset or the instances of the hard classes make up a large proportion of the dataset.

\subsubsection{Effect of class selection}
\begin{table}[h!]
    \centering
\begin{tabular}{ |c|c| } 
\hline
 Selected classes&  Detection accuracy(\%) \\
  \hline
 50 hard &   83.47   \\ 
  \hline
 50 random &  81.77  \\ 
 \hline
  70 hard & 86.85 \\
  \hline
 \end{tabular}
\\[8pt]
    \caption{Detection accuracy of easy/hard classes on CIFAR-100}
    \label{detection}
\end{table}

\begin{table}[h!]
    \centering
\begin{tabular}{ |c|c|c||c|c| } 
\hline
\multirow{2}{*}{Selected classes}& \multicolumn{2}{c|}{Training accuracy(\%)} & \multicolumn{2}{|c|}{Test accuracy(\%) } \\
 \cline{2-5}
 & main & MEANet  & main & MEANet \\
  \hline
50 hard &  69.84 & 97.22  & 50.70 &  59.76   \\ 
 \hline
50 random & 76.51 & 98.42  & 61.42 &  69.64\\ 
 \hline
70 hard & 72.70 & 94.43  & 54.53 & 57.50 \\ 
 \hline
100 & 79.20 & 92.65  & 61.70 & 62.37 \\ 
 \hline
 \end{tabular}
\\[8pt]
    \caption{Effect of selection of classes on the accuracy of selected classes of CIFAR-100 dataset, using ResNet32 A}
    \label{class_compare}
%\\[8pt]
\end{table}
 To analyze the effect of which classes to select for training, we compare selection based on class-wise complexity with random selection. In addition, we investigate the effect of the number of selected classes. Experiments are run with different numbers of selected classes including 50, 70, 100 out of 100 classes in CIFAR-100. Table.\ref{detection} shows that the accuracy of detecting whether an instance belongs to $C_{hard}$ is lower if randomized selection is used. Table.\ref{class_compare} indicates that as the number of selected classes increases, the improvement on training and test accuracy decreases although more training data are used. This is because our model benefits from the reduction in the number of classes, which makes it easier to find the decision boundaries. In conclusion, selecting based on class-wise complexity and defining a small set of hard classes are suggested to guarantee the advantage of our model.

%\begin{table}[h!]
%    \centering
%\begin{tabular}{ |c|c| } 
%\hline
%Selected classes&  Test accuracy \\
%  \hline
%50 hard &   62.20  \\ 
%  \hline
%50 random &  61.81  \\ 
% \hline
%70 hard &  62.47 \\
% \hline
% \end{tabular}
%\\[8pt]
%    \caption{Test Accuracy of CIFAR-100}
%    \label{training_compare}
%\\[8pt]
%\end{table}

\subsubsection{Training cost}
\begin{figure}
    \centering
    \includegraphics[width=2.8 in]{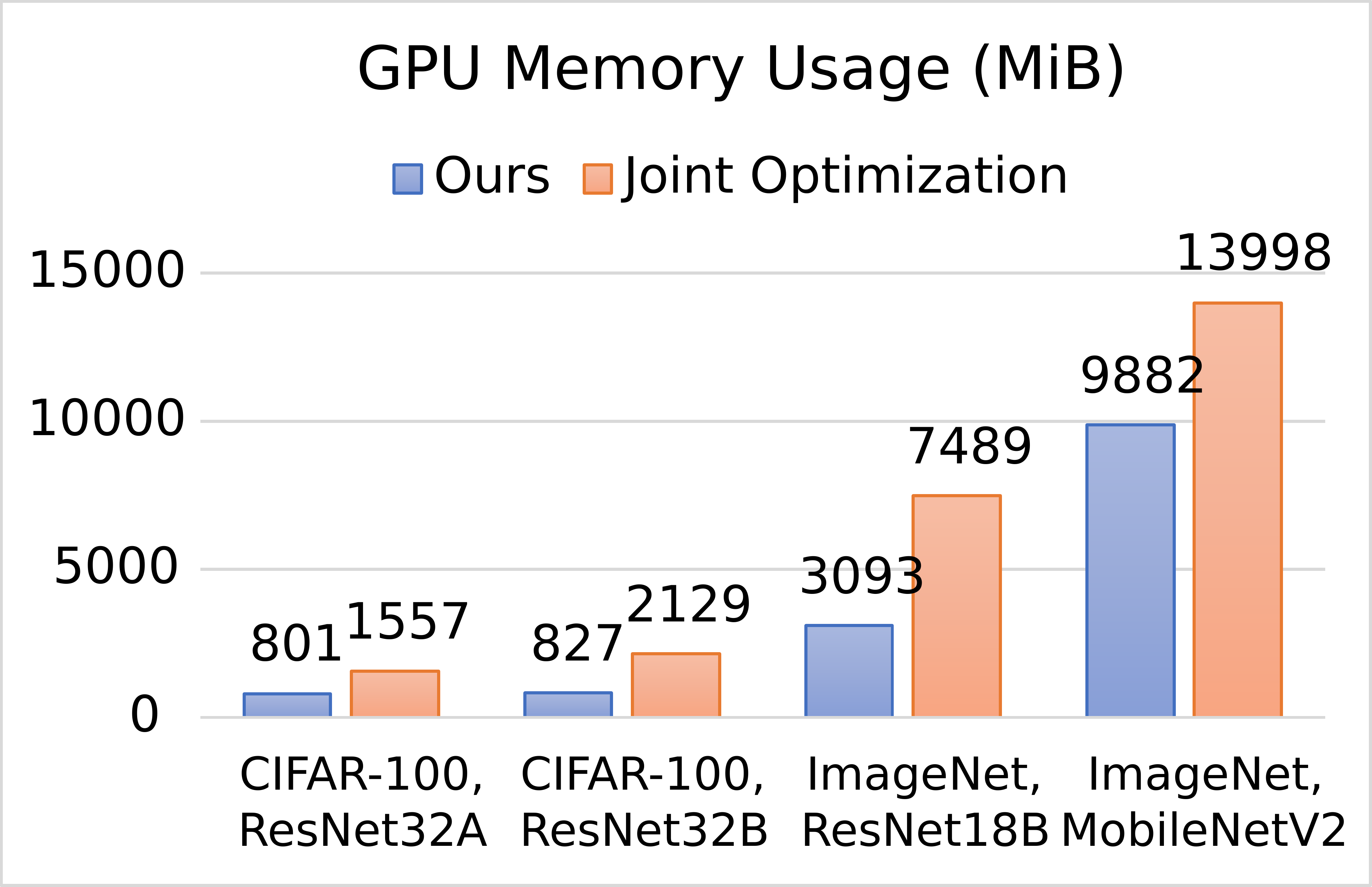}
    \caption{ GPU memory usage for training the extension and adaptive blocks with our algorithm or joint optimization (batch size = 128)}
    \label{memory}
\end{figure}

%Training energy
We then estimate the training cost of the proposed model. Because the main block can be pretrained at the cloud, we only consider the training cost of other blocks at the edge, and compare our approach with joint optimization using data of all classes. Computation cost of training is composed of forward propagation cost and backward propagation cost. Both fixed parameters and trained parameters are used in the forward pass, therefore, this part of cost is the same for our approach and the joint approach. However, fixed parameters do not require gradient computation, and they do not cause any backpropagation cost. Since our approach fix the main block, only parameters in the adaptive block and the extension block are trained. In Table.\ref{weights}, we show the numbers of parameters that are fixed or trained when using our approach. Also, we count the related number of computations in terms of Multiply-Adds using a Python package called ptflops \cite{flops}.
The number of parameters and the corresponding number of computations directly affect the memory and computation cost. While our approach uses fixed parameters only for forward propagation, joint optimization \cite{Branchy:journals/corr/abs-1709-01686} trains all the parameters together. Hence the number of computation related to fixed parameters equals to the reduced number of gradient computation of our approach compared to joint optimization. When both approaches train the model with the same batch size, our training algorithm uses 60\% less GPU memory for ResNets and 30\% less for MobileNets, as shown in Fig. \ref{memory}. % reduces the training cost per instance to 40\% and 31\% using ResNet32 A and B respectively for CIFAR100.} Similarly, the training cost per ImageNet instance is reduced to 30\% and 54\% using MobileNetV2 and Resnet18.
Furthermore, because we only use instances belonging to hard classes for training, the amount of training data is reduced by half. As a result, the total training cost to process the training dataset can be further reduced by half. %Overall, the training cost is reduced to around 20\% of the cost using joint optimization and all classes.

%Table for number of weights
\begin{table}[h!]
    \centering
\begin{tabular}{ |c|c|c||c|c| } 
\hline
\multirow{2}{*}{}& \multicolumn{2}{c||}{\# of computation} & \multicolumn{2}{|c|}{\# of parameters} \\
 \cline{2-5}
Dataset, model & fixed & trained &fixed & trained\\
 \hline
CIFAR-100, ResNet32 A & 46 & 31& 0.11 & 0.37 \\ 
 \hline
CIFAR-100, ResNet32 B & 69 & 31 &0.47 & 0.42\\ % 69, 0.47;100, 0.89
 \hline
ImageNet, MobileNetV2 B & 300 & 130 &3.49 & 1.09\\ %300, 3.5;430, 5.58
 \hline
ImageNet, ResNet18 B & 1722& 2058 & 11.16 & 27.46\\%3780-1722, 38.62-11.16 % use a deeper extension block 
 \hline
 \end{tabular}
\\[8pt]
    \caption{Number of compuatation and number of parameters (Million)}
    \label{weights}
%\\[8pt]
\end{table}

%According to Table. I, the difference is given by
%\begin{align*}
%    E_{feature} - E_{raw} = [(q-1) + (1-\beta)(\alpha_2 - \alpha_1)] Nx
%\end{align*}
%In the case of sending features, we assume that two thirds of layers are distributed at the edge and others are at the cloud, so $q=2/3$. Since the size of the output feature is 32x16x16 and the image size is 32x32, the ratio of communication energy and computation energy $\alpha_2$ for uploading features should be eight times of $\alpha_1$ for uploading images. By Table. VI, around 20\% of data are sent to the cloud, so $1-\beta= 0.2$. Then the equation becomes
%\begin{align*}
%    E_{feature} - E_{raw} = [-\dfrac{1}{3} +  1.4 \alpha_1] Nx
%\end{align*}
%If $\alpha_1> 0.24$, $E_{feature} - E_{raw}>0$. This equation also applies to latency. According to the power model of 4G and Wifi in \cite{mobilecloud}, this condition is satisfied by AlexNet and VGG16. Although $\alpha_1$ depends on devices and models, the communication power in IoT is usually comparable to the computation power. Hence, we conclude that sending raw data is faster and more energy efficient under most circumstances.

%Because latency in a IoT system is often proportional to energy and Table. I applies to both latency and energy, we skip the latency analysis and suppose that the conclusions of energy analysis hold for latency. 

\subsubsection{Distributed inference by edge and cloud}
\begin{figure}
    \centering
    \includegraphics[width=2.8 in]{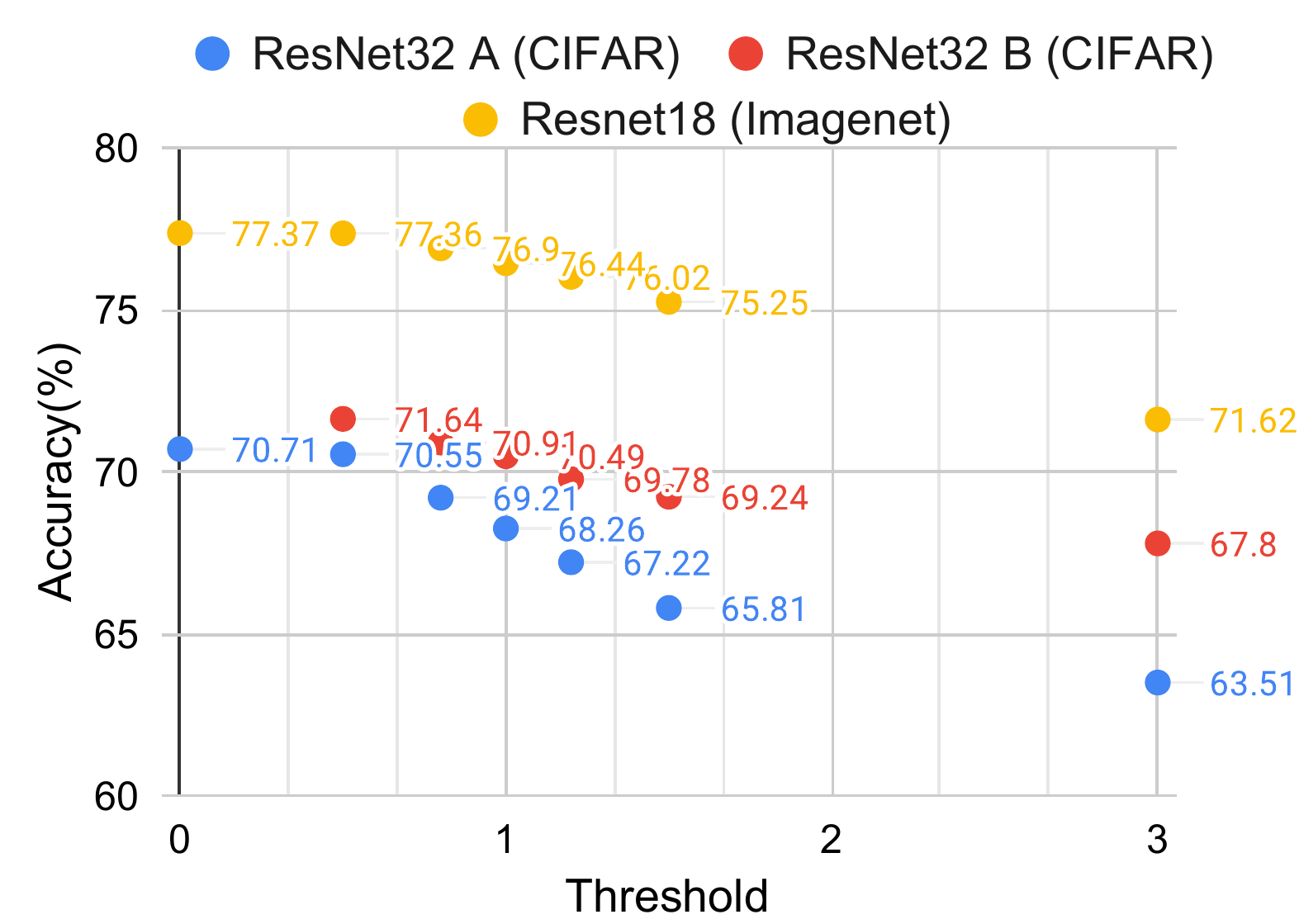}
    \\[8pt]
    \includegraphics[width=2.8 in]{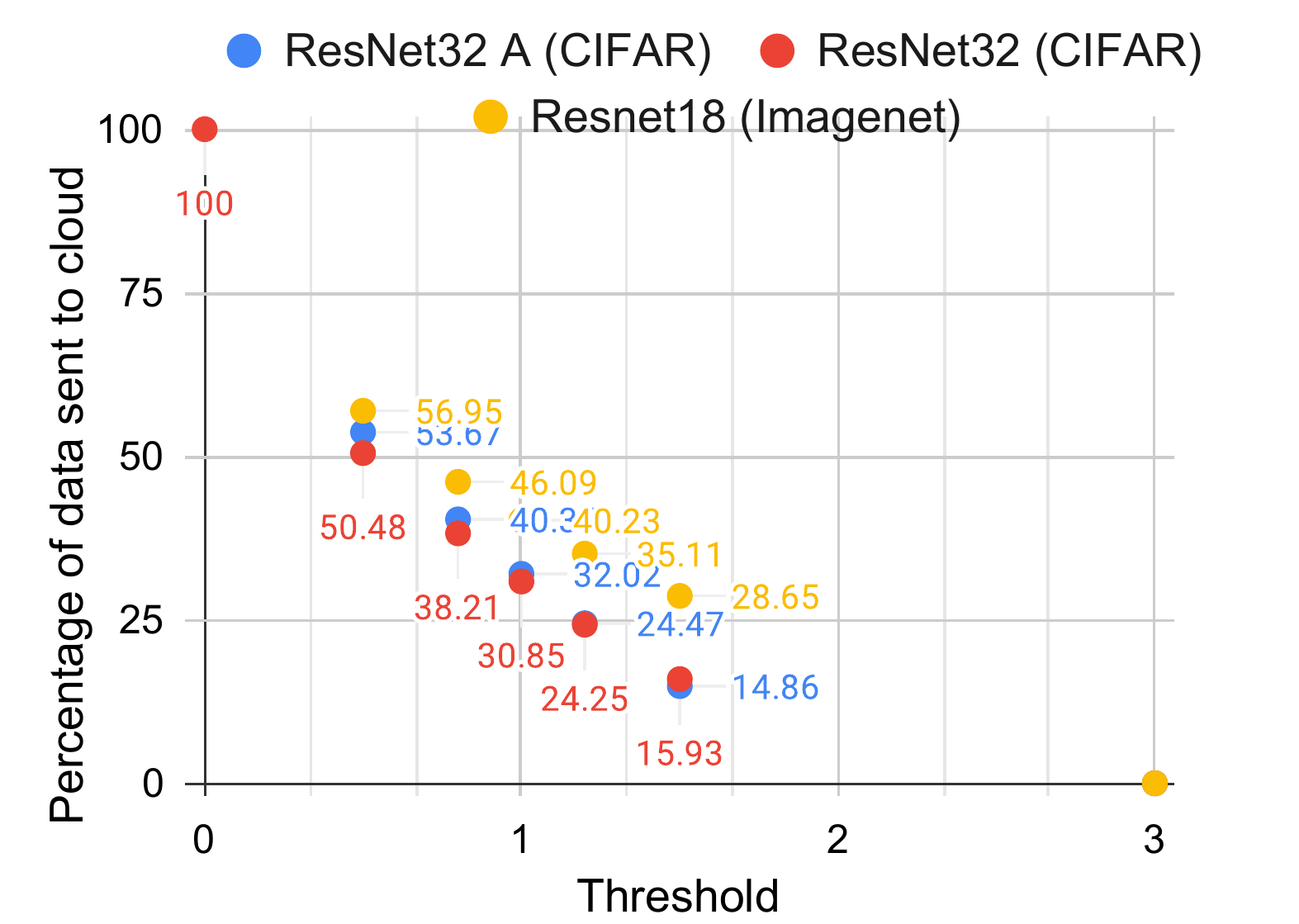}
    \caption{Distributed inference by edge and cloud}
    \label{fig:accuracy}
\end{figure}
We compare the accuracy between edge-only, cloud-only, and edge-cloud AI systems. Since the cloud server does not have a resource constraint, we use ResNet101 as the cloud AI. In Fig.\ref{fig:accuracy}, it shows the overall accuracy and the percentage of data sent to the cloud depending on the threshold. When the threshold is zero, all data are sent to the cloud. Compared to the edge-only inference, distributed inference can improve the accuracy on CIFAR-100 by 2\% by sending 15\% of data to the cloud. The accuracy improvement on ImageNet is 4\% when sending 28\% of data. If a lower threshold is set, the percentage of data sent to the cloud and the accuracy would both increase. When the threshold is low enough (0.5), distributed inference can achieve similar accuracy as using only the cloud. 
\begin{figure}%
    \centering
    \includegraphics[width=2.7 in]{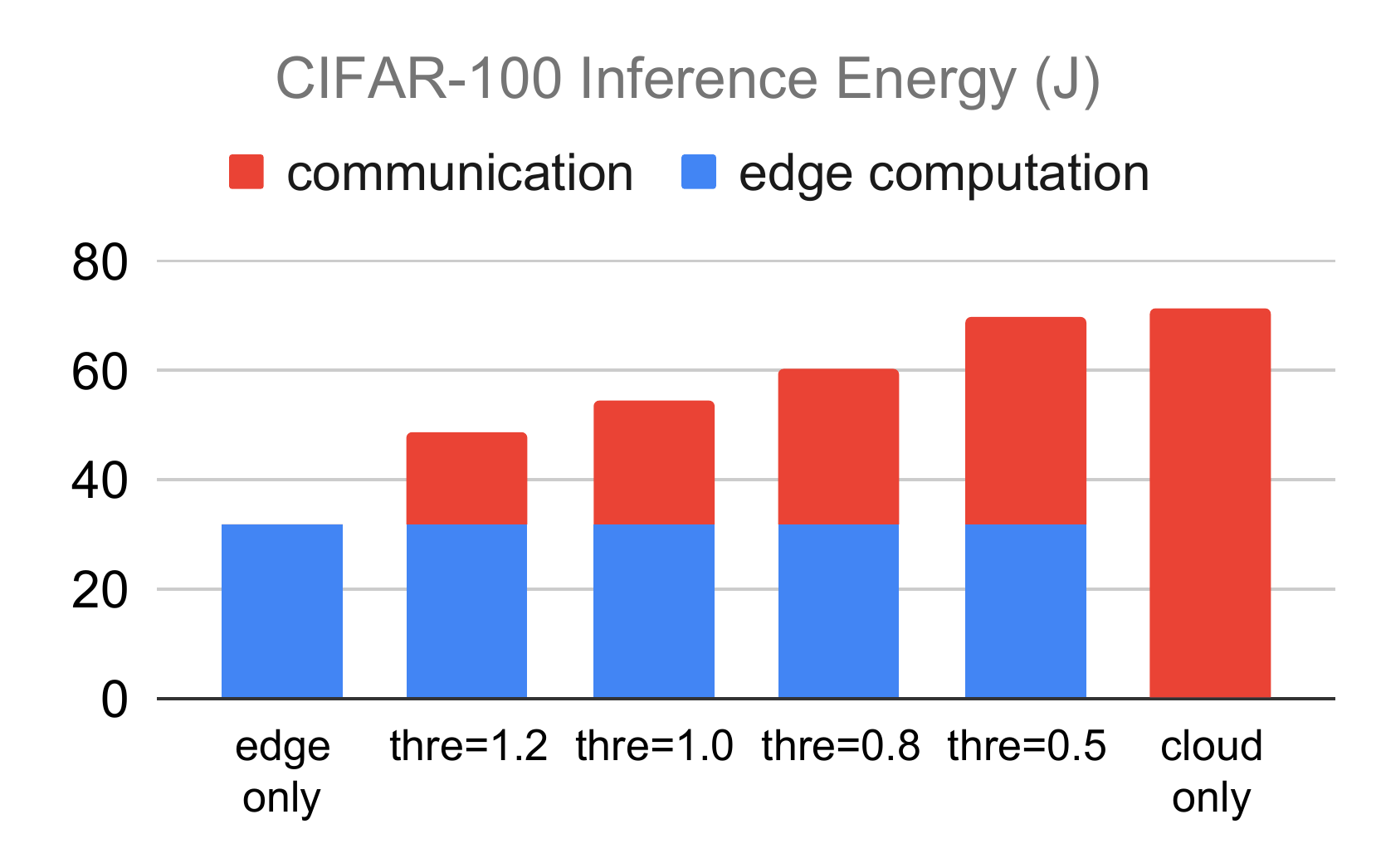}
    %\includegraphics[trim=15 0 0 10, clip,width=1.7 in]{CIFAR-100 Inference Energy.pdf}}}%
    %\qquad
    \includegraphics[width=2.7 in]{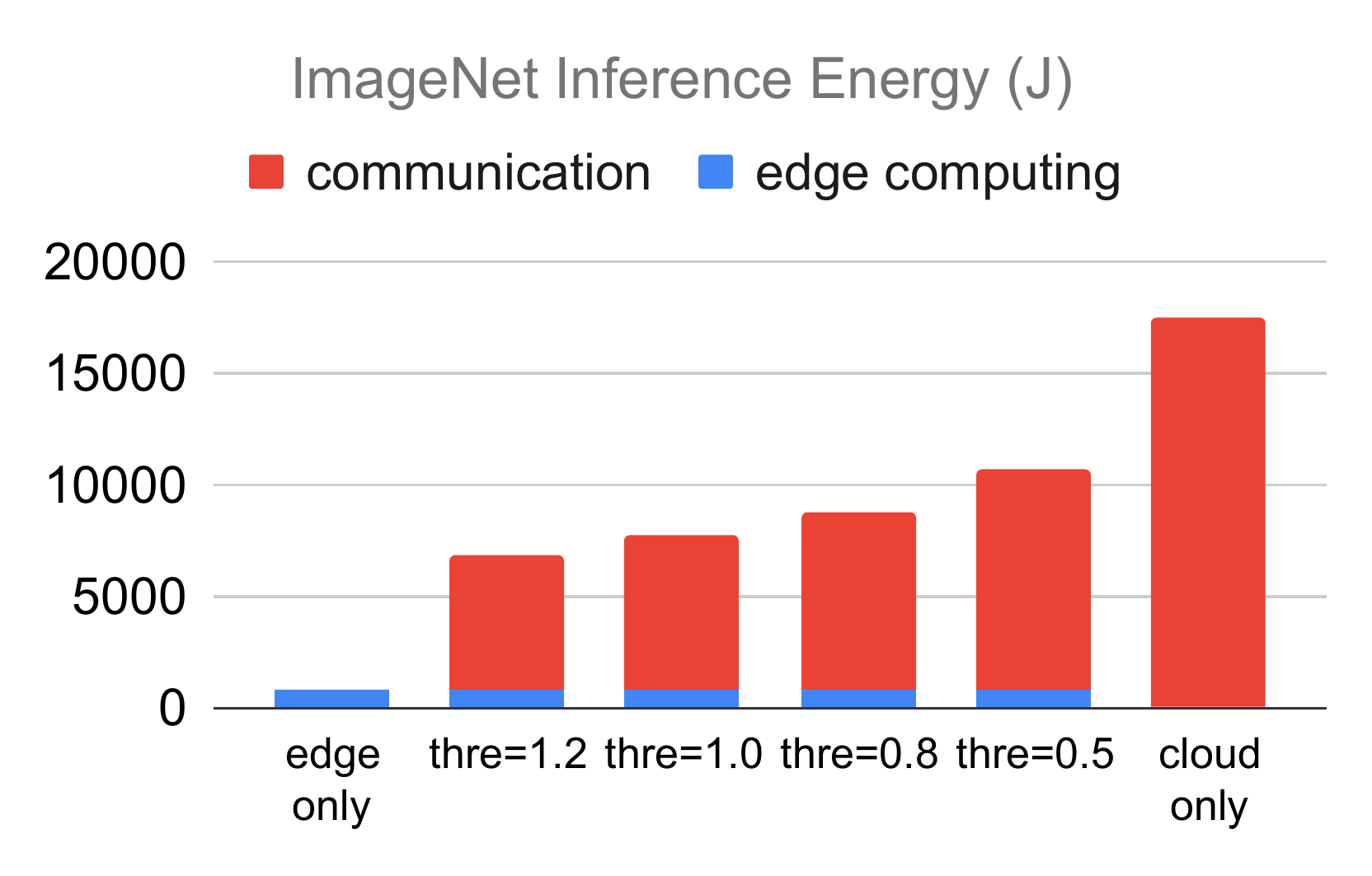}
    \caption{Total energy consumption at the edge to infer 10000 images in CIFAR-100 testset using ResNet32 A; total energy consumption at the edge to infer 50000 images in ImageNet testset using ResNet18 B}%
    \label{totalenergy}%
\end{figure}

%Inference energy 
  Next, to compare the inference energy between edge-only, cloud-only, and edge-cloud AI systems, we calculate the energy consumption at edge, which is the sum of computational and communication energy of the edge. The computation energy of the cloud AI is ignored because it is not a concern. 
 First, we calculate the communication energy to upload an image. Based on the power model of WiFi in \cite{power} and \cite{mobilecloud}, the power of uploading data is calculated by $P_{upload}= 283.17(mW/Mbps) \times s_{upload} + 132.86(mW)$, where $s_{upload}$ is the throughput. We assume that the throughput equals to the average upload speed 18.88Mb/s. Then communication power is $P_{upload}= 5.48W$. Since the image size in CIFAR-100 is 32x32x3 (bytes) and the image size in ImageNet is 224x224x3 (bytes), the communication time $t_{cu}$ to upload an image in CIFAR-100 or ImageNet is 1.3 ms or 63.7 ms respectively. Then the communication energy is calculated by $E_{cu}=P_{upload}*t_{cu}$. 
 
 Second, we estimate the computation energy and latency of the neural network at the edge. The GPU power is monitored by Nvidia system monitor. The latency for inferring an instance is the latency of GPU running a batch of instances divided by the batch size. Then the inference energy is calculated and listed in Table.\ref{edgecloud}. Please note that the GPU we use for simulating an edge-cloud system is Nvidia GeForce GTX 1080 Ti. This GPU is not designed for the edge. In a real edge-cloud system, a less powerful GPU can run the neural network at the edge with smaller batch size and may have longer latency and lower power. %However, the energy of running the same neural network using different GPUs should not have a large difference.}

 In Fig.\ref{totalenergy}, we show the total energy consumed by inferring all images in the test set of CIFAR-100 and ImageNet. We insert numbers to equations in Table. \ref{edge-cloud-compare} to calculate the total energy. It shows that when the threshold is 0.5 for CIFAR-100, the total energy consumption at the edge is close to that of sending all samples to the cloud. However, since more than 50\% of data inference have terminated at the edge, edge-cloud distributed inference still has the advantage in latency. In the case of ImageNet, the computation energy at the edge is much smaller than the communication energy because of the large image size. Hence, the distributed inference achieves the same ImageNet accuracy as the cloud by consuming only 60\% of energy at the edge.
 
 %TODO: imagenet results

\begin{table*}
    \centering
\begin{tabular}{ |c|c|c|c|c|c|c|} 
\hline
Dataset, model & GPU power(W) & WIFI uploading power (W)& $t_{cp}$ (ms)&$t_{cu}$(ms) &$E_{cp}$(mJ) &  $E_{cu}$(mJ) \\
 \hline
CIFAR-100 ResNet32 A & 56 & 5.48 & 0.056 & 1.3 & 3.14 & 7.12 \\ 
 \hline
ImageNet ResNet18 B & 75 & 5.48 & 0.203 & 63.7 & 15.23 & 349 \\ 
 \hline
 \end{tabular}
\\[8pt]
    \caption{Computation and communication power, time, and energy at the edge per image}
    \label{edgecloud}
%\\[8pt]
\end{table*}

\section{Conclusion}
%% Please rewrite conclusions. Read the abstract and Intro and write as -- We propose....the advantages and disadvantages... some critical results.
%% Please read the abstract and rewrite the abstract in your own words. The current conclusion does not read well.  We propose a distributed network architecture that lends to energy efficient and fast classification of easy instances on the edge device while going to the cloud on rare occasions when we detect hard classes. The hard/easy class identification with high confidence is done on the edge using an adaptive .....  

With energy and resource constraints at the edge, smart IoT systems are facing great difficulty to process data promptly and accurately. In this work, we propose a distributed system that lends to energy-efficient and fast classification of most instances on the edge device, while activating the cloud for complex ones on rare occasions. The complexity of an instance is evaluated by the edge using the proposed MEANet. The energy efficiency of the edge is due to conditional inference of easy and hard classes using early exits at the main or the extension block. Instances that can not be identified as easy or hard classes by the main block are considered ``complex" and sent to the cloud.
The proposed architecture, training and inference techniques are applicable to typical CNNs such as ResNets and MobileNets. The experimental results show that the proposed model can obtain higher accuracy at a low training cost, which confirms its potential for application to real-life edge devices. 
With edge-cloud collaboration, the system achieves a better accuracy-vs-energy tradeoff compared to cloud-only or edge-only approaches.
%we demonstrated that deploying trained AI models is not able to adapt to IoT environments, and cloud-only and edge-only approaches are not optimal with regard to performance and energy. 
 % A new network architecture consisting of three blocks with different functionalities is designed for the edge device. %By training the extension block and adaptive block with data of hard classes, the network has a stronger ability to distinguish between hard classes. During inference, instances can exit at the main block, the extension block based on the complexity of test input.
% For future work, we encourage researchers to apply our methodologies to real-life IoT systems. Also, we want to investigate the effect of training the edge network with a separate set of complex data collected by IoT devices.
%\section{ }

%\newpage

\section*{Acknowledgement}
The research was funded in part by C-BRIC, one of six centers in JUMP, a Semiconductor Research Corporation (SRC) program sponsored by DARPA, the National Science Foundation,  and Vannevar Bush Faculty Fellowship.
%References
\bibliographystyle{ieeetr}
\bibliography{ref}

%\newpage
%\includepdf[pages=-,pagecommand={},width=\textwidth]{response.pdf}
\end{document}